\documentclass[conference]{IEEEtran}
\IEEEoverridecommandlockouts
\usepackage{cite}
\usepackage{amsmath,amssymb,amsfonts}
\usepackage{graphicx}
\usepackage{textcomp}
\usepackage{xcolor}
\usepackage{hyperref}
\def\BibTeX{{\rm B\kern-.05em{\sc i\kern-.025em b}\kern-.08em
    T\kern-.1667em\lower.7ex\hbox{E}\kern-.125emX}}
    
\usepackage{algpseudocode}
\usepackage{xcolor,colortbl}

\begin{document}
\bibliographystyle{thebibliography}

\title{Embedded event based object detection with spiking neural network\\

\thanks{This material is based upon work supported by the French national research agency (project DEEPSEE ANR-17-CE24-0036) and by the donation from Franck DIARD to the Université Côte d'Azur.}
}

\makeatletter
\newcommand{\linebreakand}{%
  \end{@IEEEauthorhalign}
  \hfill\mbox{}\par
  \mbox{}\hfill\begin{@IEEEauthorhalign}
}
\makeatother

\author{
    \IEEEauthorblockN{Jonathan Courtois}
    \IEEEauthorblockA{\textit{LEAT, Univ. Côte d'azur, CNRS} \\
    jonathan.courtois@univ-cotedazur.fr}
    \and
    \IEEEauthorblockN{Pierre-Emmanuel Novac}
    \IEEEauthorblockA{\textit{LEAT, Univ. Côte d'azur, CNRS} \\
    pierre-emmanuel.novac@univ-cotedazur.fr}
    \and
    \IEEEauthorblockN{Edgar Lemaire}
    \IEEEauthorblockA{\textit{LEAT, Univ. Côte d'azur, CNRS} \\
    edgar.lemaire@univ-cotedazur.fr}
    \linebreakand % <------------- \and with a line-break
    \IEEEauthorblockN{Alain Pegatoquet}
    \IEEEauthorblockA{\textit{LEAT, Univ. Côte d'azur, CNRS} \\
    Alain.Pegatoquet@univ-cotedazur.fr}
    \and
    \IEEEauthorblockN{Benoit Miramond}
    \IEEEauthorblockA{\textit{LEAT, Univ. Côte d'azur, CNRS} \\
    Benoit.Miramond@univ-cotedazur.fr}
}

\maketitle

% REWORK 3 DONE from here to % --- %

\begin{abstract}
%Abstract--summary of paper: 
The complexity of event-based object detection (OD) poses considerable challenges. Spiking Neural Networks (SNNs) show promising results and pave the way for efficient event-based OD.
Despite this success, the path to efficient SNNs on embedded devices remains a challenge.
This is due to the size of the networks required to accomplish the task and the ability of devices to take advantage of SNNs benefits.
Even when "edge" devices are considered, they typically use embedded GPUs that consume tens of watts.
In response to these challenges, our research introduces an embedded neuromorphic testbench that utilizes the SPiking Low-power Event-based ArchiTecture (SPLEAT) accelerator.
Using an extended version of the Qualia framework, we can train, evaluate, quantize, and deploy spiking neural networks on an FPGA implementation of SPLEAT.
We used this testbench to load a state-of-the-art SNN solution, estimate the performance loss associated with deploying the network on dedicated hardware, and run real-world event-based OD on neuromorphic hardware specifically designed for low-power spiking neural networks.
Remarkably, our embedded spiking solution, which includes a model with 1.08 million parameters, operates efficiently with 490 mJ per prediction.
\end{abstract}


\section{Introduction} \label{Introduction}

Event-based processing has several advantages for efficient embedded solutions, especially for autonomous machines. 
It is tied to the environmental variation under study and focuses processing on the useful information, but requires specialized sensors, hardware, and algorithms to maximize the benefits.
On the sensor side, for image processing applications, event-based cameras are increasingly present in the literature as they are energy efficient and have interesting temporal resolution and dynamic properties.
On the algorithm side, deep neural networks (DNN) offer impressive solutions for image processing, but generally with a significant drawback in terms of size, complexity and energy consumption, especially for Object detection (OD).
To move towards lighter solutions based on the communication method of biological neurons, Spiking Neural Networks (SNN) offer solutions that are compatible with event data processing and are energy efficient due to the nature of their operations, as they transmit information between neurons via binary and asynchronous impulses called spikes.

Although they are often more complex to train, they are still capable of handling complex tasks such as OD and, in particular, event-based OD.
In the automotive use case, this has been demonstrated by the first fully spiking event-based solution \cite{b14} on the Prophesee GEN1 dataset \cite{b25}.

Even if SNNs can be trained and tested on GPUs, their sparse communication and temporal mechanism are not fully utilized, giving them a disadvantage when evaluated on GPUs compared to FNNs.
SNNs require special hardware, called neuromorphic hardware, to fully exploit their advantages.
This neuromorphic hardware constitutes a necessary step for a fully event-based functional approach to achieve the promised efficiency, and it also paves the way for testing in real-world, constrained use cases.
In this paper, we focus on the Spiking Low-power Event-Based ArchiTecture (SPLEAT) accelerator\cite{b20}.
It is a programmable neuromorphic architecture supporting convolutional spiking neural networks whose technical maturity has already been demonstrated in the context of in-orbit observation\cite{b20}.
Thanks to the deep neural network framework Qualia \cite{bQUALIA}(formerly MicroAI \cite{b2}), we can deploy and test SNN solutions on SPLEAT for real-world use cases.

We contribute in this paper to the development of a new, affordable approach for embedded event-based processing on a dedicated low-power chip using a real-world use case, event-based OD in the automotive domain.
The goal is to overcome the challenge of deploying and evaluating SNNs on embedded neuromorphic hardware to get an idea of the performance penalty and energy efficiency of the solution.
To the best of our knowledge, this is the first embedded spiking neural network capable of performing automotive OD with real event data embedded on a low-power event-based chip.
%BM: on pourrait nous opposer Loihi : i) spikingyolo a t il été déployé sur Loihi ? ii) avez comparer le déploiement sur SPLEAT et sur Loihi ?

In section \ref{Related work} we go over all the elements required for the workflow and then in \ref{Method} we focus on the framework, network and hardware needed for this application. 
In section \ref{Experiments} we go into the details of the experiment and show and discuss the results in section \ref{Results}. Finally, section \ref{Conclusion} concludes our work and presents future perspectives.

\section{Related work} \label{Related work}
With the proliferation of autonomous machines interacting with their environment, there is an increasing demand for sensors and processing capabilities that enable efficient operation, low latency and minimal power consumption, especially in autonomous vehicles.
In this area, OD is an important application that requires high detection performance with minimal power and memory requirements to enable real-time processing in vehicles \cite{b33}.

\subsection{Event based camera}

The detection of objects in the car would not be possible without sensors to visualize the surroundings.
LIDAR and standard cameras currently dominate the field \cite{b33}, but a new sensor is currently gaining interest \cite{b10}.
Event-based data has interesting properties for this use case, with high temporal resolution, low power consumption and a high dynamic range.
Their inherent event-based nature results from independent pixels that trigger events when brightness variations are detected \cite{b9}.
Each event contains four elements: the position on the pixel grid $(x,y)$, the timestamp $t$ and the polarity of the event $p$.
To give an order of magnitude, the Prophesee event-based camera used in the GEN1 dataset achieves a temporal resolution of about $1 \mu s$ \cite{b25}, compared to a standard camera working at $60Hz$, meaning about $17ms$ time resolution.

Due to their event-based nature, event-based cameras do not output frames. 
Instead, they provide a stream of events that are highly correlated with the dynamic scenes they observe.
However, in order to use them with a formal neural network (FNN), this stream of events must be integrated within a short time span to generate event frames that are very sparse. 
Due to their dense computation, FNNs lose the advantage of using sparse event-based data, but another form of neural network is better suited to this.

\subsection{Spiking neural networks}

Spiking neural networks consist of biologically inspired neurons that transmit their information through discrete binary impulses that are triggered when their membrane potential reaches a predefined threshold over time.
These neurons, which mimic the functionality of biological neurons, also introduce a temporal dimension to their operating model that allows them to process data over time, and particularly, event-based data.

The most commonly used models of spiking neurons are the Integrated and Fire (IF) and the Leaky Integrated and Fire neurons (LIF)\cite{b4}.
IF neurons accumulate input values within their membrane potential both spatially and temporally until it reaches a threshold value.
Time is discretized into time steps for computational purposes.
For each time step, the neuron calculates its neuronal charge, compares it to the threshold, emits a spike if necessary, and then updates its internal state.
LIF neurons use a similar methodology to IF neurons. 
They introduce a leak mechanism during the neuronal charge step.
The leak factor is set manually via a time constant $\tau$ or can be learned during training with Parametric Leaky integrate and fire neuron (PLIF).

Due to the event-based nature of spiking neurons and their information processing method, they are well suited for energy-efficient solutions, especially in resource-constrained environments and even more so when the hardware can take advantage of sparse computation.

Despite these advantages, a major challenge with spiking neurons lies in the non-differentiable nature of the threshold mechanism, which is often computed by a Heaviside activation function that prevents the gradient descent training process in neural networks.
Furthermore, the introduction of time-steps can lead to latency and potential problems with gradient vanishing during training, especially in the context of large networks or a high number of time-steps.
This time step represents the discrete integration time of the input signal in order for the network to output a prediction.
Approaches such as Truncated Backpropagation Through Time (TBPTT) and Surrogate Gradient learning (SG) \cite{bLOICthese}\cite{b1}, applied to SNNs, have emerged to address these challenges, significantly contributing to the reduction of training difficulties.
This is especially notable when employing direct encoding methods, aligning with the practices seen in conventional neural networks.

\subsection{Event Frame}

In supervised direct encoding methods, the dense, non-binary input is directly fed into the first layer of the network
However, event based data, by their temporal resolution and format, are not inherently compatible with the training methods and GPU support commonly associated with Deep Neural Networks (DNNs).
A common approach involves accumulating events over a short period to reconstruct a dense frame referred to as an 'event frame'.
This frame can be conceptualized as having two channels, one for positive events and another for negative ones \cite{b5}.
While using a sum for event accumulation results in frames with integer values \cite{b6}, an alternative approach is binary accumulation, as demonstrated in \cite{b7}.
This method maintains events in a binary frame format, leveraging the energy efficiency advantages inherent to spiking neural networks on dedicated hardware.

\subsection{Event based Object detection}

Object detection, especially for automotive applications, offers an ideal testing ground for event-based sensors and SNNs.
The task includes both the localization of objects (object localization) and the determination of their identity (object recognition) in an environment with strong fluctuations in light and motion.
To achieve this goal, a reliable, energy- and memory-efficient method is required for seamless integration into embedded solutions.
For this task, FNNs have emphasized the use of the Single Shot MultiBox Detector (SSD) \cite{b8} to efficiently map image pixels to bounding box coordinates and class probabilities, essentially treating OD as a regression and classification task.
Designed for real-time processing and optimizable in size, this OD algorithm offers promising opportunities for real-time embedded applications.

\begin{figure*}[htbp]
    \centering{\includegraphics[width=0.33\linewidth,angle=90]{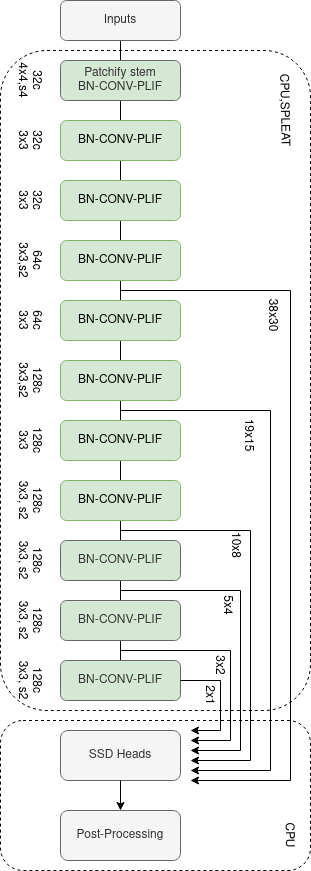}}
    \caption{ST-VGG+SSD architecture \cite{b29} with the layer distribution plan where c stands for channels, s stride and 3x3 kernel's size.}
    \label{fig1}
\end{figure*}

Event-based OD poses challenges due to the unique nature of the data. 
Despite these challenges, various approaches —formal, hybrid, or spike-based— have demonstrated the feasibility of achieving impressive performance.
The Recurrent Event-camera Detector (RED) approach presented in \cite{b10} is directly trained on event frames, utilizing temporal information to generate outputs for classes and bounding box details. 
RED, a feed-forward recurrent neural network, incorporates convolutional layers and ConvLSTM to feed Single Shot MultiBox Detector (SSD) heads with extracted features.
Another notable approach is ASTMNet, detailed in \cite{b11}, which leverages SSD heads with VGG backbone architectures.
By extracting event-based features, the authors in \cite{b12} propose a recurrent vision transformer (RVT-B) that is optimized for event streams and achieves the actual top result on the GEN1 dataset with a 47.2mAP for 18.5M parameters on a 70W Nvidia T4 GPU.
Hybrid approaches, such as the hybrid SNN \cite{b13}, integrate an SNN backbone to extract event-based features that are then fed into ANN heads.
Towards energy-efficient solutions, \cite{b14} presents the first fully spiking neural network approach where an SSD head is fed with spiking sparse features derived from a spiking backbone (VGG-11, MobileNet-64, or DenseNet121-24).
In addition, a lightweight version, SpikeThin-VGG (ST-VGG) \cite{bLOICthese}, has been developed. 
It is composed of 1.08 million parameters, while maintaining performance comparable to \cite{b14}.
Another spiking approach, EMS-Res10 \cite{b15}, uses a spiking ResNet that redefines the spiking ResNet block and processes dense features.

These researches highlight the interest of event-based OD and spike-based approaches in achieving low energy consumption for embedded solutions.
However, it is important to note that the majority of these studies provide training and testing results on GPUs. 
Yet, the final and natural evaluation of SNN should be done on dedicated embedded hardware.

\subsection{Neuromorphic Hardware}

Special hardware is required to exploit the sparsity and specificity of spike calculations.
These systems, commonly referred to as neuromorphic hardware, usually consist of a Neural Processing Unit (NPU) with specific calculation rules \cite{b22}.
The architectures are either based on analog (Neurogrid \cite{b18}) or digital electronics (Spinnaker \cite{b16}, Truenorth \cite{b17}, Loihi \cite{b19}, SPLEAT \cite{b20}). 
Numerous new platforms are under development, offering the possibility of using more or less biologically plausible neurons, but all rely on sparsity and event-based computing, demonstrating the research interest in this area.
Moreover, the studies in \cite{bICONIP} and \cite{b24} show that SNNs can be up to 8 times more efficient than FNNs when computed on dedicated digital hardware, carefully considering memory reuse and sparsity.

\section{Method} \label{Method}

In this work, we adopt the specifications of the neuromorphic SPLEAT accelerator \cite{b20, b21} to give the details on the constraints of our deployment environment. 

\subsection{SPLEAT}
The SPiking Low-power Event-based ArchiTecture is an RTL (Register Transfer Level) Soft IP designed for very-low-power SNN inference acceleration.
This accelerator targets both FPGA and ASIC implementations.
SPLEAT is written in VHDL and features fully digital Neural Processing Units (NPUs) optimized for sparse convolution processing.
In SPLEAT, each spiking layer is implemented in a dedicated NPU.<
SPLEAT NPUs work in an event-based fashion, i.e. they become active when they receive input stimuli in the form of spikes.
Thanks to this feature, the NPUs can benefit from SNNs sparse activity and thus achieve significant energy savings.
%Thanks to this feature, the NPUs can benefit from \st{the spike sparsity in} SNNs \textcolor{red}{sparse activity} and thus achieve significant energy savings.
SPLEAT supports IF, LIF and PLIF models for spiking neurons.
It also supports different layer configurations: fully connected layers, convolutional layers and pooling layers.
Thanks to the batch norm fusion provided by \cite{bLOICthese}, SPLEAT is also able to support batch normalisation layers.
These layers can run with user-defined fixed point dynamics (from 2 bits to 64 bits), which can be different at the layer level for weights, activations and potentials.
The hardware limitations of SPLEAT are bound to the available FPGA resources of the targeted board for implementation.

\subsection{SpikeThin-VGG}

In \cite{b14} 3 spiking neural network models were presented and trained with the Gen1 Automotive Detection dataset \cite{b25}.
This dataset contains 39 hours of event-based road situations recorded with a resolution of 304x204, 
which corresponds to the GEN1 camera. 
The camera was mounted on the dashboard of a car.

It contains 228,000 bounding boxes for cars and 28,000 bounding boxes for pedestrians.

The performance evaluation of the OD model is based on the COCO mAP metric \cite{b26}, 
which represents the mean average precision over 10 intersection over union (IoU), with thresholds from 0.50 to 0.95.
\cite{b14} established the best achievable COCO mAP of 0.189 for SNN on this dataset thanks to a neural architecture gathering both a DenseNet121-24 backbone and a SSD model.
Inheriting this research, we developed the SpikeThin-VGG (ST-VGG) \cite{bLOICthese} in order to targets highly constrained embedded applications.
The ST-VGG consists of a patchify stem input convolution \cite{bLOICthese}, seven convolutional blocks (CONV) in the backbone and another 3 additional CONV layers, each ending with a PLIF activation.
During training, each CONV block contains a CONV layer preceded by a batch normalization (BN) layer and followed by the PLIF.
From these, six spiking features are extracted to feed the SSD Head (see Fig. 1).
The output of the regression and classification SSD heads are post-processed to obtain bounding boxes of the detected element and its predicted class.
The particular placement of the BN and CONV layers was justified as it led to better results and had no impact on the fusion process.
However, the BN block introduces real values that disrupt the binary information flow within the network.
As shown in \cite{b29}, the BN-CONV block can be fused to a single CONV' after training, preserving the binary information flow during inference.
Furthermore, ST-VGG was trained by continuous learning with hard reset for spiking neurons over an entire video clip of 60s.
In this approach, the entire video clip is considered as a sample consisting of 1200 $50ms$ event frames.
Each event frame is presented to the network once. The network outputs predictions for each event frame.
During the sample, the potential of each PLIF is reset to 0 if it needs to trigger a spike.
At the end of the 1200 event frame, the entire potentials of the network are reset to 0 and move on to the next video clip.
In this way, the gradient graph is unrolled across all event frames and TBPTT can be used when a ground truth occurs on a part of the gradient graph to save memory and avoid gradient vanishing, as explained in section 5.3.1 of \cite{bLOICthese}.
With this configuration, the network achieves a COCO MaP of 0.184 \cite{b26} on the Gen1 Automotive Detection dataset \cite{b25} for an SNN with $1.14M$ parameters consisting mainly of convolutional layers and resulting in a lighter and easier configuration for our embedded perspective.

\subsection{Qualia}

We use the Qualia framework\cite{bQUALIA} to create our testbench.
Qualia is an end-to-end framework designed for the computation and deployment of deep neural networks.
With its testbench configuration file, we can control various steps in the workflow, including training, evaluation, quantization and also deployment on microcontrollers for FNN and SNN and on FPGA boards with SPLEAT configuration for SNN only.

In this work, we will only use the deployment part of Qualia, as there is a plugin that allows us to generate C code from a Spikingjelly \cite{b3} Python network architecture and the subsequent VHDL configuration for an FPGA SPLEAT deployment.
For compatibility reasons, we define the C spiking neurons based on the Spikingjelly framework dynamic.
The framework can convert IF, LIF and PLIF from Python to C code.
The generated C code ensures that we have control over the parameters and activation arithmetic.
Before the conversion, Post-Training Quantization (PTQ) can be applied to the parameters and activation.

Furthermore, thanks to Qualia's ability to generate and test different network configurations, we can estimate performance degradation, plan the deployment and simulate the results without having to implement the network on the FPGA.
Given the time required for synthesis and implementation, this method proves to be a valuable time- and resource-saving strategy.

\section{Experiments} \label{Experiments}

In the present results, SPLEAT was configured with a 16-bit fixed-point format in a Q8,8 arithmetic (8bit for the integer part including sign, 8 for the decimal part) for all layers.
For our experiments we used an AMD Zynq UltraScale+ MPSoC ZCU102 board with an XCZU9EG FPGA with a 100~MHz clock for the programmable logic and an embedded quad-core ARM Cortex-A53 CPU running at 1200~MHz.
A comprehensive overview of dedicated FPGA hardware for SNN processing is provided in \cite{b32}.
The authors mainly focus on classification tasks on the MNIST dataset with a wide variety of architectures and results ranging from $2.5k$ to $1.861k$ synapses consuming between $59.09W$ and $0.18W$. These results makes a comparison difficult due to the differences in datasets, network architectures and task complexity.
To enable a better comparison with a simpler dataset, we tested SPLEAT for two applications.
One on the Google Speech Commands V2 dataset with a four-layer spiking convolutional neural network, described in \cite{bICONIP}.
The other on the Prophesee Gen1 dataset \cite{b25} with a variant of the 32-ST-VGG network \cite{bLOICthese}.
\begin{table}[htb]
\caption{Detail on the deployed solutions on SPLEAT with a XCZU9EG FPGA. *Only the backbone.} 
\label{tab1}
\resizebox{\columnwidth}{!}{%
\begin{tabular}{|
>{\columncolor[HTML]{C0C0C0}}c |cc|}
\hline
\textbf{Dataset}      & \multicolumn{1}{c|}{\cellcolor[HTML]{C0C0C0}\ \ GSC V2 - 35 cls\ } & \cellcolor[HTML]{C0C0C0}Prophesee Gen1 \\ \hline
\textbf{Neurons bitwidth}                & \multicolumn{1}{c|}{16}     & 16       \\ \hline
\textbf{Weights bitwidth}                & \multicolumn{1}{c|}{16}     & 16       \\ \hline
\textbf{Model}                     & \multicolumn{1}{c|}{LIF}    & LIF      \\ \hline
\textbf{Available BRAM}               & \multicolumn{2}{c|}{912}               \\ \hline
\textbf{Used BRAM}                 & \multicolumn{1}{c|}{22.5}   & 740      \\ \hline
\textbf{Available DSP}                & \multicolumn{2}{c|}{2520}              \\ \hline
\textbf{Used DSP}                  & \multicolumn{1}{c|}{22}     & 61       \\ \hline
\textbf{Available Logic Cells}        & \multicolumn{2}{c|}{274080}            \\ \hline
\textbf{Used Logic Cells}          & \multicolumn{1}{c|}{7537}   & 27415    \\ \hline
\textbf{Architecture} & \multicolumn{1}{c|}{SCNN 4-Layers}                           & small 32-ST-VGG*                       \\ \hline
\textbf{Synapse}                   & \multicolumn{1}{c|}{25763}  & 886752   \\ \hline
\textbf{Latency / output {[}ms{]}} & \multicolumn{1}{c|}{1}      & 700      \\ \hline
\textbf{Power {[}W{]}}             & \multicolumn{1}{c|}{0,2}    & 0,7      \\ \hline
\textbf{Energy / output {[}mJ{]}}  & \multicolumn{1}{c|}{0,2}    & 490      \\ \hline
\textbf{Energy / synapse {[}nj{]}} & \multicolumn{1}{c|}{7,76}   & 553      \\ \hline
\textbf{Performance}                  & \multicolumn{1}{c|}{92.8\% accuracy} & 14.4 mAP \\ \hline
\end{tabular}%
}
\end{table}
\subsection{Google Speech Commands V2}
As a simple use case, we first tested a small spiking convolutional neural network with the Google Speech Commands V2 dataset \cite{b27}.
This dataset contains audio clips with 35 words and a duration of 1 second each.
As explained in \cite{bICONIP}, the audio clips were converted to spectrograms using MFCC, resulting in an input of 48x10 dimensions.
The network with 2 time steps is fed with the input, which is divided into two parts with temporal dependence (2x24x10).
The input is treated as a 1D image with 2x24 samples and 10 channels.
In this way, the network treats a different part of the spectrograms at each time step and outputs the classification of the corresponding word after these two time steps.
This network consists of four convolutional layers with the respective $xc$channels, $x$features and $sx$ stride of [48c3s1 - 48c3s1 - 96c3s1 - 35c1s1].
Each convolutional layer is followed by a batch normalization layer (which is fused after training) and LIF activation with a leakage factor of 2.

To be used with SPLEAT, this $25k$ parameter network, trained in float arithmetic, had to be quantized to conform to SPLEAT fixed-point arithmetic.
The $P$ parameters were converted to Q8,8 by multiplying them by $2^8$ and then rounding downwards to a 16-bit integer.
For spiking neurons, the values for threshold, leakage and membrane potential must also be quantized.
Thanks to this small size and this arithmetic conversion, this network fits easily on the FPGA.

\subsection{Event-based object detection}
The realistic use case was OD in the event-based dataset Prophesee Gen1 \cite{b25}, which was treated with a variant of the 32-ST-VGG network \cite{bLOICthese}.
With its $941k$ backbone parameters, the 32-ST-VGG network fills the BRAM capacity of the FPGA board used, making its use impossible.
The Small 32-ST-VGG is a lighter version of the 32-ST-VGG network with smaller features in the backbone \ref{fig1}, reducing the backbone parameters to $886k$, resulting in a network with $1.08M$ parameters including the SSD head and post-processing.
The lighter backbone (green cell in Figure \ref{fig1}) and the extra layers consist of convolutional layers with the respective $xc$ channels, $x$ features and $sx$ stride of [32c4x4s4, 32c3x3s1, 32c3x3s1, 64c3x3s2, 64c3x3s1, 128c3x3s2, 128c3x3s1, 128c3x3s2, 128c3x3s2, 128c3x3s2, 128c3x3s2].
Six spike feature maps of size [38x30, 19x15, 10x8, 5x4, 3x2, 2x1] are extracted and passed to the classification and regression SSD heads.
The SSD heads consist of convolutional layers and post-processing is used to filter the predictions generated by the predictor heads with the anchors.
Furthermore, PLIF are implemented as LIF, where the learned value for leakage is constant for the inference.

Using such a network on SPLEAT requires customization, as the SSD heads and post-processing are not supported by SPLEAT.
This prompted us to proceed in two ways:
We test the entire network on the embedded CPU and then test the partitioning of the network between the embedded CPU and SPLEAT for spiking event-based processing.
In this planning we distinguish 3 parts of the network: the backbone (including the extra layers), the SSD heads, and the post-processing.
The SSD heads and post-processing are always implemented on a CPU (embedded or workstation).
The backbone and the extra layers are implemented on SPLEAT or CPU depending on the tested configuration, as shown in Figure \ref{fig1}.
While the implementation on SPLEAT requires fixed-point arithmetic, we can use floating point or fixed point on the embedded CPU.

Since the entire data set does not fit on the board, we will stream the event frames to the board.
Our test method consists of a client-server communication, where the server executes the neural network and the client sends the $50ms$ event frames of the video clip and receives the positions of the bounding boxes and the classification.
The COCO maP calculation is performed after all predictions have been received.

\begin{table*}
    \caption{Inference detail of the small 32-ST-VGG for the entire Gen1 Dataset.}
    \centering
    \resizebox{2\columnwidth}{!}{%
    \begin{tabular}{c|c|c|c|c|c|c|c|c|c|c}
    \hline
    \rowcolor[HTML]{C0C0C0} 
    \textbf{Name} &
      \textbf{Backbone target} &
      \textbf{Arithmetic} &
      \textbf{Param} &
      \textbf{Spike / output} &
      \textbf{Activity} &
      \textbf{maP} &
      \textbf{Time / output} &
      \textbf{Bckb time} &
      \textbf{SSD time} &
      \textbf{Postproc time} \\ \hline
    \rowcolor[HTML]{FFFFFF} 
    ref 32-ST-VGG\cite{bLOICthese}      & Workstation GPU & float & 1,14 M & 377,5k    & 36,24\% & 0,184 & x        & x         & x       & x      \\ \hline
    \rowcolor[HTML]{FFFFFF} 
    small 32-ST-VGG     & Workstation CPU & float & 1,08M  & 214,9k  & 32,05\% & 0,144 & 150ms & 128ms  & 20ms & 2ms \\ \hline
    \rowcolor[HTML]{FFFFFF} 
    small 32-ST-VGG     & Workstation CPU & Q8,8 & 1,08M  & 215,1k  & 32,08\% & 0,144 & 219ms & 197 ms & 20ms & 2ms \\ \hline
    \end{tabular}%
    }
    \label{tab2}
\end{table*}

\begin{table*}
    \caption{Inference detail of the deployed small 32-ST-VGG for a single sample of the Gen1 Dataset. *: executed on Embedded CPU.}
    \centering
    \resizebox{2\columnwidth}{!}{%
    \begin{tabular}{c|c|c|c|c|c|c|c|c}
    \hline
    \rowcolor[HTML]{C0C0C0} 
      \textbf{Backbone target} &
      \textbf{Arithmetic} &
      \textbf{Spike / output} &
      \textbf{Activity} &
      \textbf{maP} &
      \textbf{Time / output} &
      \textbf{bckb time} &
      \textbf{SSD time} &
      \textbf{postproc time} \\ \hline
    \rowcolor[HTML]{FFFFFF} 
    \cellcolor[HTML]{FFFFFF}Workstation CPU & float & 213,5k    & 31,84\%  & 0,321 & 143ms      & 122ms  & 19ms  & 2ms        \\ \hline
    \rowcolor[HTML]{FFFFFF} 
    Workstation CPU                         & Q8,8 & 213,7k     & 31,87\%  & 0,323 & 207ms      & 186ms  & 19ms  & 2ms        \\ \hline
    \rowcolor[HTML]{FFFFFF} 
    \cellcolor[HTML]{FFFFFF}Embedded CPU    & float & 213,5k    & 31,84\%  & 0,321 & 1697ms     & 1486ms & 202ms & 9ms        \\ \hline
    \rowcolor[HTML]{FFFFFF} 
    Embedded CPU                            & Q8,8 & 213,7k     & 31,87\%  & 0,323 & 1818ms     & 1543ms & 266ms & 9ms        \\ \hline
    \rowcolor[HTML]{FFFFFF} 
    \cellcolor[HTML]{FFFFFF}SPLEAT          & Q8,8 & 214,8k     & 32.04\%      & 0,337 & 975ms      & 700ms  & 266ms* & 9ms*             \\ \hline
    \end{tabular}%
    }
    \label{tab3}
\end{table*}

\section{Results} \label{Results}
The main constraint for FPGA (or SoC) integration remains the memory embedded in the circuit, over and above the constraints of logic resources or DSP blocks.
Most state-of-the-art networks do not allow such integration as discussed in section \ref{Related work}. Our efforts to provide a model offering the best compromise between memory footprint and performance have resulted in the small 32-ST-VGG.\\
As it can be observed from the table \ref{tab2}, the 32-ST-VGG baseline \cite{bLOICthese} achieves a COCO 0.184 mAP with 1.14M parameters for the whole test set, while the small 32-ST-VGG implemented in C with float arithmetic achieves a 0.144 mAP for 1.08M parameters.
We observe a COCO mAP loss of 0.04 for a difference of 60k parameters.
Interestingly, the parameter reduction also results in a decrease of the average spike accumulation per output (with 162.6k fewer spikes on average), and the average activity of the network decreased by 4\% from 36.24\% to 32.05\%.

After quantization, the small 32-ST-VGG, with a 16-bit fixed-point backbone, achieves a similar mAP as the float network on the entire test set.
The notable differences are that the backbone inference time, both on the workstation and the embedded CPU, is higher for the fixed-point version, which could be partly due to the conversion of the extracted spike feature maps from fixed-point to float for the SSD heads, which is included in this time.
This result also shows that the 16b fixed-point arithmetic does not lead to performance degradation.
The execution of the small 32-ST-VGG on the test set was performed on a workstation equipped with 2$\times$ Intel Xeon Gold 6230R CPUs, however, only a single core was used for the entire prediction.

However, switching from the workstation CPU to the embedded ARM Cortex A53 CPU leads to a significant difference in the inference time.
as can be seen from the tables \ref{tab2} and \ref{tab3}. 
The workstation CPU requires an average of 150 ms to process a $50ms$ event frame and provide its prediction, while the embedded CPU requires an average of 1700 ms. 
These differences, which are mainly due to the lower clock speed and lower performance micro-architecture  of the embedded CPU, increase the total duration of the experiments from a few hours on the workstation (thanks to multi-threading across samples) to more than a week on the embedded CPU.

To save resources and time, we decided to continue the experiment with a single sample, as shown in Table \ref{tab3}.
As expected, only the inference time is affected. When switching from the workstation to the embedded CPU, the activity and performance remain similar.
However, when comparing the embedded CPU version with Q8,8 arithmetic for the backbone and the SPLEAT version, we find that the performance of the network is maintained with a backbone time for inference reducing from 1818 ms to 975 ms.

Figure \ref{fig3} illustrates the differences in the predicted bounding boxes with different methods for a representative video clip.
Here, a red bounding box shows the network with a float arithmetic for the backbone on the workstation CPU with an mAP of 0.321, followed by the 16-bit quantized backbone on the workstation CPU in green with 0.323 mAP and the SPLEAT output in blue with an mAP of 0.337.

Even with a time gain of 46.3\%, the network outputs a prediction after 975ms, which is too much given the input of 50ms and therefore prevents the application to process data in real-time for now.

Compared to the results of the GSC v2 with the 4-layer SCNN, where the network outputs a prediction after two time steps in 1ms, the small 32-ST-VGG backbone outputs features maps in 700ms feeding the SSD heads and post-processing that output predicted bounding boxes 275 ms later reaching 975 ms

This backbone latency overhead is largely due to the low parallelism of the SPLEAT architecture: 
the network is implemented as a pipeline of consecutive NPUs, with a single NPU per layer.
The different layers are therefore processed in parallel.
In a particular layer, however, the spikes are processed sequentially by a single NPU.
After receiving a spike, the NPU calculates the membrane potential updates for all stimulated neurons in sequence, one neuron at a time.
Therefore, the computation latency for a single spike strongly depends on the size and the number of kernels involved in the convolutional layers.
Therefore, networks with a high number of spikes and a large size and number of kernels are at a disadvantage on SPLEAT.
The small 32-ST-VGG suffers from a 4.37 times higher number of kernels than the 4-layer SCNN, which also transitions from 1D to 2D kernels as can be seen in Table \ref{tab4}.
In addition, the activity of the network was not optimized during training, resulting in an activity of 32.05\% for the entire network, which generated over 214.8k spikes to produce an output prediction, which is 30 times more than the 4-layer SCNN.
This augmentation adds additional latency due to communication time and spike address calculation, feature map extraction and conversion to CPU.
Combining the number of spikes and the kernel, the larger architecture requires 130 times more kernel calculations for an inference.
If we look at the 11 spiking layers of the backbone in detail (see table \ref{fig2}), we can see that the number of spikes and the activity vary greatly from the input to the last layer.
This table illustrates the high sparsity of inputs, which reduces their involvement in the latency problem, but also shows potential bottleneck layers that can paralyze the entire network.
To solve this problem, we are currently working on an improved version of SPLEAT that provides parallelism at the layer level.
By using multiple NPUs for a single layer and distributing the computational load evenly across them, we expect a strong acceleration.
Initial measurements on toy examples show a speedup by a factor of 4 when using 16 NPUs instead of 4 for a 4-layer CNN on the Fashion MNIST dataset.
It is also important to note that parallelism in SPLEAT does not increase the memory footprint, which is the main limiting factor in neuromorphic systems.
However, the use of logic cells and DSPs increases proportionally to the number of NPUs in the design.

\begin{figure}[t]
    \centering{\includegraphics[width=1\linewidth]{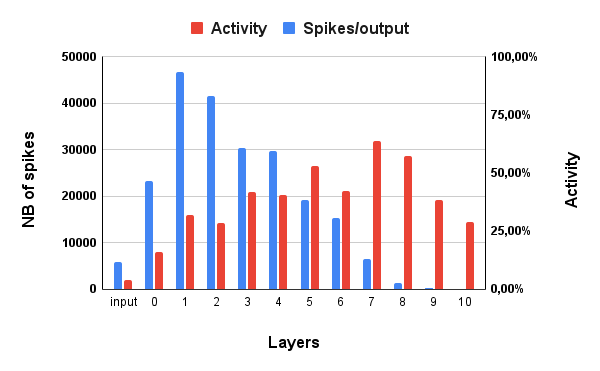}}
    \caption{Mean spike accumulation and activity per output per video clip for the 16 bit quantized Backbone network on workstation CPU with a total mean of 215076 spikes per video clip and 32.05\% activity.}
    \label{fig2}
\end{figure}

This enlargement of the neuronal architecture leads to both an increase in latency and in energy.
Table \ref{tab1} shows estimations of the utilization of logic resources (logic cells, block RAMs, DSPs...) and the power consumption measured for SPLEAT.
These results were obtained through post-implementation estimations using the AMD Xilinx toolchain.
The power consumption estimation is based on a post-implementation simulation, which strengthen the reliability of the estimated value.
Compared to the 4-layer SCNN, the small 32-ST-VGG backbone increases its dynamic power consumption from 200mW to 700mW.
This increase of 3.5 follows from the increase of 3.6 logic cells used for this network.
Combined with the latency, this results in an energy consumption of 0.2 mJ for the SCNN, which scales up to 490 mJ for the small 32-ST-VGG backbone.
Table \ref{tab4} summarises the keypoint ratio between the two networks, where the energy metrics corresponds to SPLEAT consumption. The Energy$_{spike}$ corresponds to the energy cost of a spike and the Energy$_{norm}$ to the normalized energy per synapse and per time step.
The Energy$_{norm}$, is more representative from the deployed architecture, independently from its activity. 
Otherwise, the Energy$_{spike}$ is directly linked to the network activity
\begin{table}
    \caption{Scaling comparison between the two networks: SCNN for keyword spotting and 32-ST-VGG for Object Detection. The numbers total spikes, latency and energy are provided per inference.}
    \resizebox{\columnwidth}{!}{%
    \begin{tabular}{c|c|c|c}
            \cellcolor[HTML]{CCCCCC} & \cellcolor[HTML]{CCCCCC} Ratio & \cellcolor[HTML]{CCCCCC} SCNN & \cellcolor[HTML]{CCCCCC}Small 32-ST-VGG \\ \hline
    Inputs               & 608         & 240     & 145920      \\ \hline
    Time steps           & 0,5         & 2       & 1           \\ \hline
    Activity             & 2,21        & 14\%    & 32\%        \\ \hline
    Synapse              & 34,42       & 25763   & 886752      \\ \hline
    Total spikes         & 29.83       & 7200    & 214800      \\ \hline
    Kernels              & 4.37        & 227     & 992         \\ \hline
    Latency [ms]         & 700         & 1       & 700         \\ \hline
    Power [W]            & 3,5         & 0,2     & 0,7         \\ \hline
    Energy [mJ]          & 2450        & 0,2     & 490         \\ \hline
    Energy$_{spike}$ [$\mu$J]  & 80        & 0,028    & 2,30        \\ \hline
    Energy$_{norm}$  [$\mu$J]  & 142       & 0,004   & 0,553       \\ \hline
    \end{tabular}%
    }
    \label{tab4}
\end{table}
With this information on SPLEAT consumption, it is possible to estimate the total energy cost of the OD solution both in software and hardware.
To obtain an estimate for the embedded CPU, the UltraScalePlus\_XPE\_2023 estimator gives an average power consumption of 1.8 W for the embedded CPU.
After measuring the latency of the overall solution on the embedded CPU, the full-software solution is estimated to consume about 3.3 J per output with an inference time of 1818 ms.
On the other side, when accelerating the backbone on SPLEAT, the SSD head and post-processing cause additional costs of about 0.5 J per output with their inference time of 275 ms.
The additional cost of post-processing can be optimized by converting it to fixed-point arithmetic to avoid conversions and take advantage of the specific optimizations for the target instruction set architecture.
This part must indeed remain on the embedded CPU since it is tailored to a specific task (here the post-processing of OD).
With a backbone energy consumption of 490mJ, the total energy cost of the hardware/software solution is about 1J per output, which means an energy gain by a factor of 3 thanks to the accelerator.
Several software and hardware optimizations are already planned in order to further improve these gains in the near future, as explained in the Conclusion and Perspectives section.

\section{Conclusion and perspectives}
\label{Conclusion}
Our study has demonstrated the successful deployment of a real use case on a low-power FPGA board, confirming the effectiveness of the Qualia - SPLEAT spiking event-based workflow for embedded solutions.
The neuromorphic SPLEAT architecture has interesting scaling possibilities and is competitive for both simple and complex applications.
The updated version of the Qualia framework now enables the use of Spiking Neural Networks (SNN) on FPGA platforms using our Spiking Accelerator.
This upgrade enables a more comprehensive understanding of the performance impact before the actual FPGA deployment, so that preventative measures can be taken to mitigate losses and effectively optimize resource utilization.

This paper presented the first embedded SNN solution for automotive event-based object detection.
Using the SPLEAT accelerator in combination with an embedded microcontroller enables 46.3\% faster inference, resulting in three times less energy consumption than software-only execution, with a loss of 0.04 mAP compared to our SNN references.
Thus, this first embedded OD solution stays within a total budget of 3.1 MB of memory at a total power of 1 W and executes in less than 1 second.

However, this advance opens up opportunities for future research to further minimize deployment losses and improve the overall performance and latency of our embedded Spiking Neural Networks. It also enables the targeted exploration of techniques aimed at addressing this problem following a codesign methodology that jointly optimizes the model and the underlying neuromorphic architecture.
As future work, SPLEAT development will therefore continue to accelerate execution time.
A short-term perspective is to support residual layers to cover a wider variety of networks and also enable parallelization of spiking layers across multiple NPUs.
This next version of our software-hardware framework would make it possible to achieve the best trade-off between real-time processing and low power consumption, thus moving closer to autonomous near-sensor processing and opening up more use cases.

\begin{figure}[htbp]
    \centering{\includegraphics[width=1\linewidth]{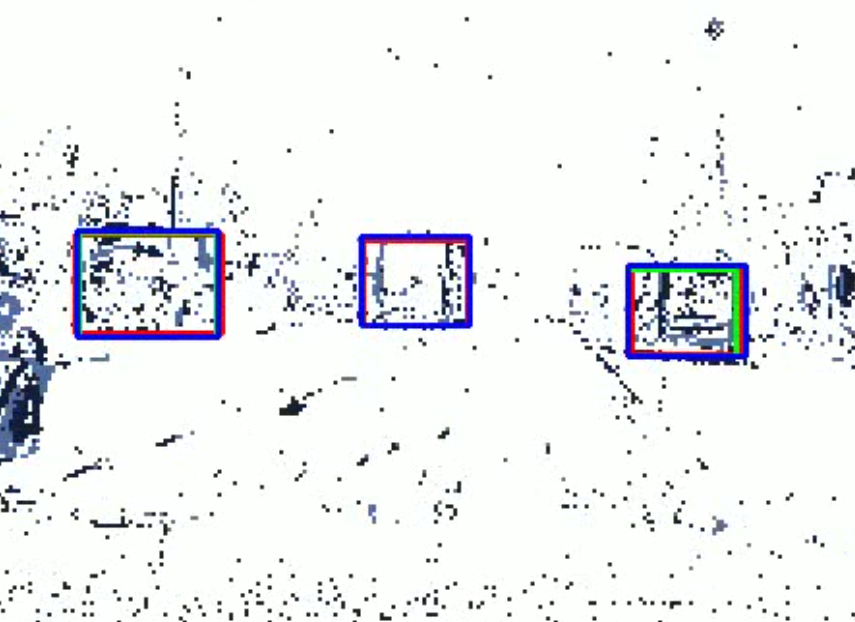}}
    \caption{Output bounding boxes of the small 32-ST-VGG with its backbone on $CPU_{float}$ (red), $CPU_{16 bits}$ (green), $SPLEAT$ (blue). $50$ms event frame extracted from the 1200 event-frames of the 60s video clip : \url{https://leat.univ-cotedazur.fr/edge}}
    \label{fig3}
\end{figure}

\end{document}